\documentclass{article}

\usepackage[nonatbib, final]{nips_2016}

\usepackage{graphicx}
\usepackage{mathtools,xparse}
\usepackage{enumitem}
\usepackage{multirow}
\usepackage{rotating}
\usepackage{mathtools,xparse}
\usepackage[ruled]{algorithm2e}

\usepackage[utf8]{inputenc} 
\usepackage[T1]{fontenc}    
\usepackage{hyperref}       
\usepackage{url}            
\usepackage{booktabs}       
\usepackage{amsfonts}       
\usepackage{amsmath}
\usepackage{nicefrac}       
\usepackage{microtype}      

\title{Learning to Drive using Inverse Reinforcement Learning and Deep Q-Networks}

\author{
  Sahand Sharifzadeh$^1$\quad Ioannis Chiotellis$^1$\quad Rudolph Triebel$^{1,2}$\quad Daniel Cremers$^1$\\
  $^1$ Department of Computer Science, Technical University of Munich, Germany\\
  \texttt{\{sharifza, chiotell, triebel, cremers\}@in.tum.de} \\
  $^2$ Institute of Robotics and Mechatronics, Department of Perception and Cognition \\
    German Aerospace Center (DLR), Oberpfaffenhofen-Weßling, Germany \\
    \texttt{ rudolph.triebel@dlr.de}
}

\begin{document}

\maketitle

\begin{abstract}
  We propose an inverse reinforcement learning (IRL) approach using Deep Q-Networks to extract the rewards in problems with large state spaces. We evaluate the performance of this approach in a simulation-based autonomous driving scenario. Our results resemble the intuitive relation between the reward function and readings of distance sensors mounted at different poses on the car.
  We also show that, after a few learning rounds, our simulated agent generates collision-free motions and performs human-like lane change behaviour.
  
\end{abstract}
\section{Introduction}
Robots and autonomous systems are becoming more and more a part of everyday life by assisting us in various different tasks. Therefore one important requirement for these systems is that they behave in a human-acceptable, socially normative way, e.g. by respecting personal spaces or treating groups based on the social relations of the individuals \cite{kruse2013human,triebel15spencer}.  This also means that humans should not just be regarded as obstacles and that an optimal robot motion must also consider human comfort metrics~\cite{okallearning}. 
One of the most popular instances of autonomous systems in the current decade is self-driving cars. The ultimate goal of autonomous cars is to drive the passengers from one point to another without any human input, while assuring the comfort of human passengers. Defining ``comfort'' is not straight forward and this makes it hard to define a suitable objective function for motion planning. One widely used method aiming to fulfil this objective is introduced by Werling et al.~\cite{werling2010optimal}. They propose to provide ``ease and comfort'' by producing jerk-optimal trajectories. Later, lane change experiments conducted by Tehrani et al., in Japanese highways showed that the human lane change behavior cannot be modelled by a single stage of jerk-optimal trajectories~\cite{tehrani2014evaluating,tehrani2015general}. Instead they proposed a two-stage model. Many other models exist similar to these. However, they mostly fail to produce human-like behaviors. Modeling the human driving behavior becomes even more complicated when considering scenarios such as driving in large cities with many intersections, traffic lights, pedestrians, etc. Therefore, applying machine learning methods to extract the models directly from the expert demonstrations appears more promising.

In a recent work, Bojarski et al.~\cite{bojarski2016end}, proposed an end-to-end supervised learning approach that maps the front facing camera images of a car to steering angles, given expert data. However, they require a large amount of data from different possible driving scenarios in order to give a good approximation of the policy. Still, they might fail when facing scenarios that are very different from the ones in the training data. A more promising formulation for this problem is based on Markov Decision Processes (MDPs). In this framework, one can apply Inverse Reinforcement Learning (IRL) to extract the unknown reward function of the driving behavior~\cite{ng2000algorithms}. By approximating this function rather than directly learning the state-action pairs in a supervised fashion, one can handle new scenarios better.
Finding the reward function of an MDP using IRL is proposed by Ng and Russel~\cite{ng2000algorithms} and further improved by Abbeel and Ng~\cite{abbeel2004apprenticeship}. Since then, several variations of IRL have been proposed such as Bayesian IRL~\cite{ramachandran2007bayesian}, maximum entropy based methods~\cite{ziebart2008maximum} and max margin prediction~\cite{ratliff2006maximum}. Most of the recent methods have been inspired by these works. For example, Wulfmeier et al.~\cite{wulfmeier2015deep} proposed a maximum entropy based approach that handles nonlinear reward functions using deep neural networks. However, most of these approaches are limited to small state spaces that cannot fully describe real-world driving scenarios. One of the main reasons is the difficulty of applying the Reinforcement Learning (RL) step in large state spaces. While approximating the Q-function for large state spaces has been effectively addressed by Mnih et al., using a Deep Q-Network (DQN)~\cite{mnih2015human}, to the best of our knowledge, it has not been used in IRL methods before.

In this paper, we address the exploding state space problem and build upon the projection-based IRL method by~\cite{abbeel2004apprenticeship} using a DQN. We implemented a highway driving simulator and evaluated the performance of our approach by analyzing the extracted rewards. The evaluation is presented in the Section \ref{eval}.
\section{Problem Formulation}
A Markov Decision Process (MDP) is defined as the tuple $(S,A,T,\gamma,R)$, where $S$ and $A$ are the state and action spaces, $T$ is the transition matrix, $\gamma \in [0,1]$ is the discount factor and $R: S \times A \to \rm I\!R$ is the reward function. A \textit{policy} $\pi: S \to A$ maps a state to an action. It can also be defined as a probability distribution over actions at each state. A \textit{value} or \textit{state-value} function $V^{\pi}(s_{0})$ is defined as the expected discounted future reward if we start from initial state $s_{0}$ and act according to policy $\pi$:
\begin{equation}\label{valuefunceq}
V^{\pi}(s_{0}) = \mathbb{E}[\Sigma_{t}\gamma^{t}R(s_t)|\pi]
\end{equation}
The discount factor applies the amount of uncertainty that we have about the future rewards.
The action-value function $Q$ represents the value that we can gain if we start from state $s_{0}$ and take action $a$, thereby applying policy $\pi$:
\begin{equation}
Q^\pi(s_0,a) = \mathbb{E}[\Sigma_{t}\gamma^t R(s_t)|\pi, a]
\end{equation}
The motion planning problem can be formulated as a Markov Decision Process, in which finding the optimal action-value function, sometimes referred to as Q-function, is the goal. Given an unknown MDP, except for the reward function, Q-Learning methods can find the optimal Q-function~\cite{watkins1989learning}\cite{watkins1992q}. Approximating the Q-function can help extend these methods to problems with larger state spaces. Using neural networks for this purpose had been shown to cause instabilities or divergence\cite{tsitsiklis1997analysis} in the past. In order to address these problems, Mnih et al., proposed two key ideas \cite{mnih2015human}. First,
to randomly sample training data from the sequence of past experiences (``experience replay''). Second, to separately train a copy of the Q-network using predictions of the original network and only periodically replace the original network with the copy.

In motion planning, similar to many other real-world applications, the MDP, including the reward function, is unknown. Inverse Reinforcement Learning methods have been shown to effectively find the underlying reward function given expert demonstrations, when the MDP is small and known, except for the reward function~\cite{ng2000algorithms}~\cite{ziebart2008maximum}~\cite{ramachandran2007bayesian}.
These approaches are mainly based on an iterative refinement of the reward function followed by a Reinforcement Learning step. In order to apply such approaches to large state spaces, we propose using a Deep Q-Network~\cite{mnih2015human} as the Reinforcement Learning step. Here, we apply this approach to the projection-based IRL method~\cite{abbeel2004apprenticeship} but other IRL techniques can also benefit from it.

Given expert demonstrations, we want to generate policies $\pi$ whose values are close to the value of the expert policies $\pi_E$:
\begin{equation}
\lVert V^{\pi}(s_0) - V^{\pi_E}(s_0)\rVert \leq \epsilon
\end{equation}
Every state $s_i$ is spanned by $d$-dimensional feature vectors $\phi(s_i)$ such as speed, acceleration, sensor readings, etc. The reward function is defined as a weighted linear combination of these features:
\begin{equation}
\label{eq:reward}R(s_i) = w \cdot \phi(s_i)\ \text{,}
\end{equation}
where $w \in \rm I\!R^d$ is the weight vector and $\lVert w \rVert \leq 1$. Plugging \eqref{eq:reward} into \eqref{valuefunceq}, we get
\begin{equation}
V^{\pi}(s_{0}) = w \cdot \mathbb{E}[\Sigma_{t}\gamma^{t}\phi(s_t)|\pi]
\end{equation}
Furthermore, feature expectations are defined as:
\begin{equation}
\mu(\pi) = \mathbb{E}[\Sigma_{t}\gamma^{t}\phi(s_t)|\pi]\ \text{.}
\end{equation}
Thus, the problem is reduced to generating trajectories whose feature expectations are similar to those of the expert. Abbeel $\&$ Ng proposed an iterative projection-based method to solve it~\cite{abbeel2004apprenticeship}. In this paper, we propose to use a DQN in the RL step of their algorithm. The details of our method are given in Algorithm \ref{algdeepprojection}.

\begin{algorithm}[t]
1. Randomly initialize the parameters of DQN and some policy $\pi^{(0)}$, compute or approximate its features expectations $\mu^{(0)}$, and set $w^{(1)} = \mu_E - \mu^{(0)}$, $\bar{\mu}^{(0)} = \mu^{(0)}$ and $i = 2$.\newline

2. Set $\bar{\mu}^{(i - 1)} = \bar{\mu}^{(i - 2)} + \frac{(\mu^{(i - 1)} - \bar{\mu}^{(i - 2)})^T(\mu_E - \bar{\mu}^{(i - 2)})}{(\mu^{(i - 1)} - \bar{\mu}^{(i - 2)})^T(\mu^{(i-1)} - \bar{\mu}^{(i - 2)})}(\mu^{(i-1)} - \bar{\mu}^{(i - 2)})$\newline

3. Set $w^{(i)} = \mu_E - \bar{\mu}^{(i - 1)}$ and $t^{(i)} =  \lVert w \rVert_2$.\newline

4. If $t^{(i)} \leq \epsilon$, terminate \newline

5. Update the DQN using $R = (w^{(i)})^T\phi$ and compute the optimal policy $\pi^{(i)}$. \newline

6. Compute or estimate features expectations $\mu^{(i)}$ of the newly extracted policy.\newline

7. Set $i = i + 1$ and go to step 2.\newline
\caption{Projection-based IRL using DQN}\label{algdeepprojection}
\end{algorithm}
\section{Evaluation}\label{eval}
\begin{figure}[h]
  \centering
  \includegraphics[scale=0.3]{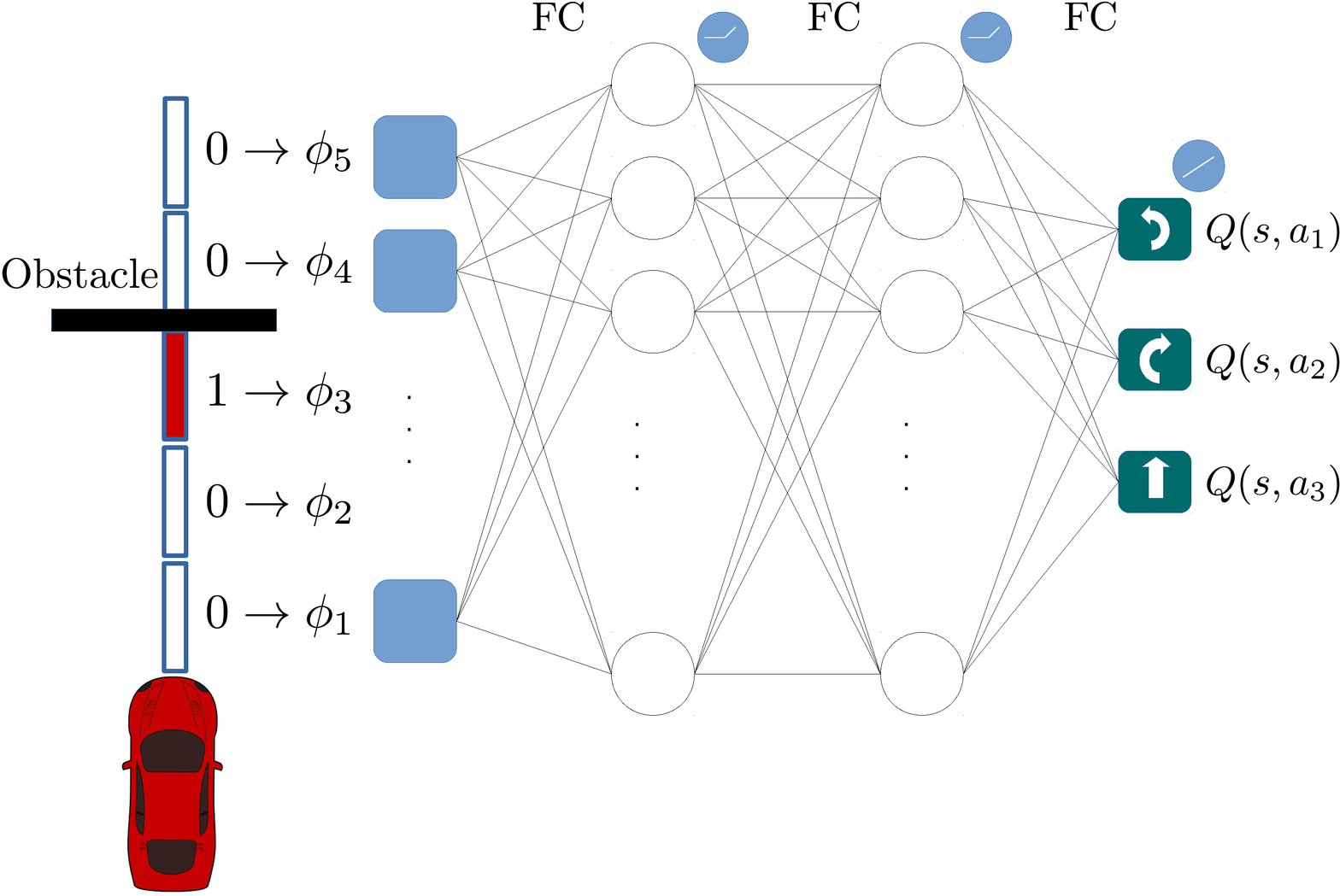}
  \caption{The architecture of the proposed Deep Q-Network. The input is the set of features and the output layer consists of action-values for three possible actions, to steer left, steer right or not steer.}\label{network}
\end{figure}

In this section, we present the evaluation results of the proposed approach. We considered the driving scenario in a highway and implemented a simulator for collecting expert trajectories and testing. The simulating environment was programmed in Python. The user interface is shown in Figure \ref{colorcodes} on the right side. The red car is the agent being trained to drive. The dynamics of this car are implemented based on the single track model~\cite{lavalle2006planning}, with three degrees of freedom. The Deep Q-network architecture used in our approach is shown in Figure \ref{network}. It consists of an input layer of features, 2 fully connected hidden layers with 160 units each and rectifying nonlinear units, followed by a fully connected output layer to the actions. 

In these experiments we used 13 sensors. The sensors had a maximum sensing radius 64$\%$ of the environment length, discretized to 16 bins of equal size. Each feature $\phi_k$, indicated whether or not there was an obstacle in the interval of each bin. If the sensor was not sensing any obstacles, the maximum possible reading distance was assigned to it. Therefore, we had a total of 208 binary features which gave rise to $2^{52}$ possible states.

In the driving experiments by~\cite{abbeel2004apprenticeship}, the car could only have discrete transitions to the lane on its left or right. However, in our experiments, we allowed steering with three different angles ($0, \frac{\pi}{12}, -\frac{\pi}{12} $) giving rise to more realistic, continuous transitions. For simplification in our experiments we set the acceleration to zero. The highway in our experiments had 3 lanes and at most two other cars could appear in front of the agent. For training, we collected 90 expert demonstrations from this setup.

The algorithm achieved satisfactory results after only 6 IRL iterations with 3000 inner loop iterations each. Since we did not have access to the true reward function of driving, we evaluated our proposed algorithm in the following ways:
\begin{enumerate}[leftmargin=0cm,itemindent=.5cm,labelwidth=\itemindent,labelsep=0cm,align=left,label=\textbf{\alph*}.]
\item\textbf{Interpreting the extracted weights:}
The extracted weights for features from 7 sensors are plotted in Figure \ref{colorcodes}. The color-code guide of each sensor is shown in the upper right corner. As shown in this figure, there is a nonlinear relation between readings of the same sensor and their extracted weights. This means that if the sensor readings had not been discretized into binary features, the algorithm would not have been able to capture the weights correctly.
\begin{figure}[h]
  \centering
  \includegraphics[scale=0.36]{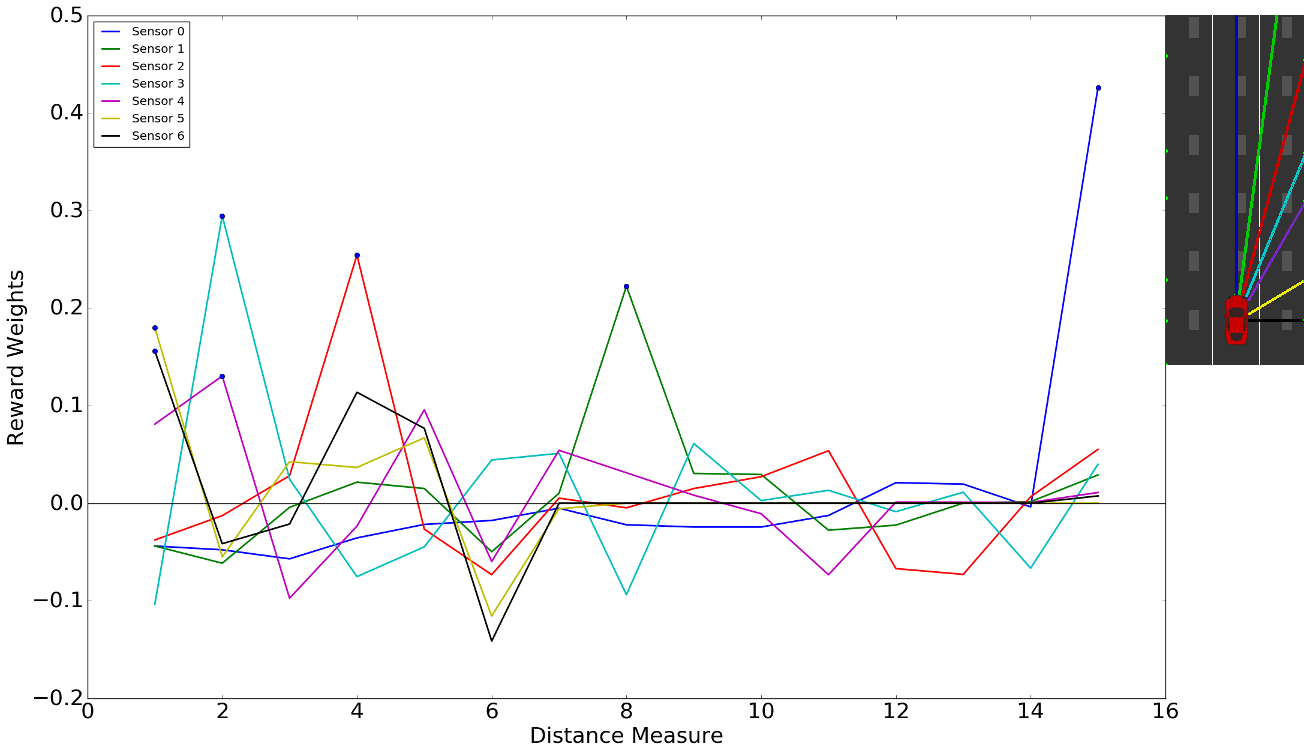}
  \caption{Weights extracted using the IRL algorithm plotted for features of 7 sensors. Each sensor has been assigned a color which is shown in the right corner.}\label{colorcodes}
\end{figure}
We have represented the maximum weight of features for each sensor with blue dots. As one can see, for Sensor 0 (blue), the higher the reading is, the larger the weight is. Meaning, the agent learns to keep as far away from the obstacles as possible. In the optimal state, where the car stays as far away from all obstacles and walls, the sensor readings form an ellipse. Therefore, as the angle of the sensor from the vertical axis increases, the reward weights peak at a smaller distance.

Another notable observation is that Sensor 6 (side sensor), gets the highest weight when reading the smallest distance from the obstacles. The explanation is that at this distance, the car is placed next to a highway wall. This has been the expert's preference in the demonstrations. This reward weight immediately drops when the reading is increased, which perfectly explains that staying in the lanes is preferable to driving between the lanes. The same can be observed for most of the other sensors. Note that some of the distances were never read by the sensors during the experiments and the extracted weights for these cases is 0. For example Sensors 5 and 6 never see obstacles in distances higher than  the width of the highway.

\item\textbf{Comparing feature expectation values of the expert to the trained agent:} We trained the DQN using the rewards computed by the final weights. Then, we let the agent drive for several scenarios, acting according to the policy of the trained network. The mean difference over feature expectations in these scenarios was computed. Part of these values are presented in Table \ref{tableResults}.

\item\textbf{Evaluating using classical motion planning objectives:}
Classical motion planning methods are evaluated based on their performance in obstacle avoidance, jerk optimality, driving in the lanes, etc. In our experiments, similar to the expert, the agent avoided the walls and obstacles in $100\%$ of the scenarios, while maintaining its position in the lane except during obstacle avoidance. Jerk values were also found to be close to the expert's. Figure \ref{scenarios} demonstrates the agent's planned motion based on the rewards extracted at different stages of the training phase.
\begin{figure}
	\centering
	\begin{tabular}{c c c c}
		\includegraphics[scale=0.43]{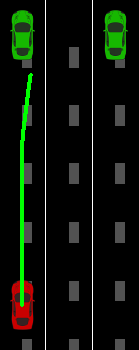}
		& \qquad
		\includegraphics[scale=0.43]{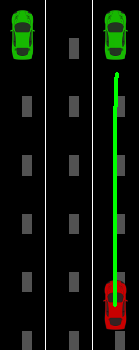}
		& \qquad
		\includegraphics[scale=0.43]{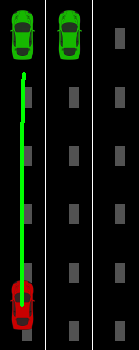}
		& \qquad
		\includegraphics[scale=0.43]{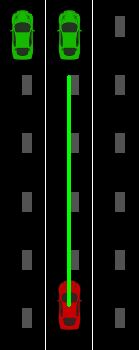}\\
		\small(a) & \qquad \small(b) & \qquad \small(c) & \qquad \small(d)
	\end{tabular}
	
	\begin{tabular}{c c c c}
		\includegraphics[scale=0.43]{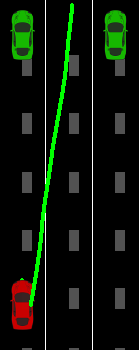}
		& \qquad
		\includegraphics[scale=0.43]{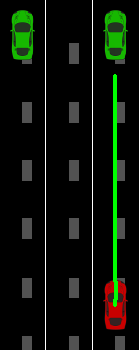}
		& \qquad
		\includegraphics[scale=0.43]{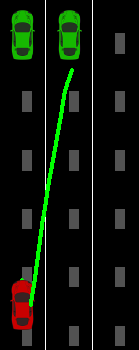}
		& \qquad
		\includegraphics[scale=0.43]{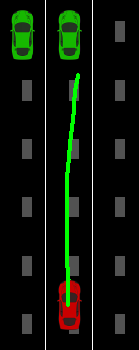}\\
		\small(e) & \qquad \small(f) & \qquad \small(g) & \qquad \small(h)
	\end{tabular}
	
	\begin{tabular}{c c c c}
		\includegraphics[scale=0.43]{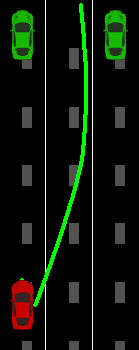}
		& \qquad
		\includegraphics[scale=0.43]{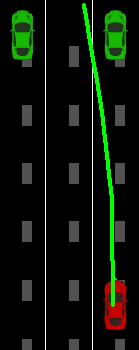}
		& \qquad
		\includegraphics[scale=0.43]{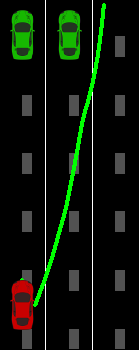}
		& \qquad
		\includegraphics[scale=0.43]{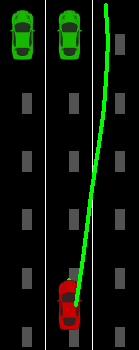}\\
		\small(i) & \qquad \small(j) & \qquad \small(k) & \qquad \small(l)
	\end{tabular}
	\caption{The agent's motion when facing four particular scenarios during training. The top row depicts the planned motion after the first IRL iteration (with 3000 DQN inner iterations), the middle row after four and the bottom row after 6 IRL iterations.}\label{scenarios}
\end{figure}
\begin{table}[h!]
\centering
\begin{tabular}{ |c c|c|c|c|c|c|c|c| } 
 \toprule
     && \multicolumn{7}{c|}{Sensor}\\
	&& 0 & 1 & 2 & 3 & 4 & 5 & 6 \\ \toprule
\multirow{7}{*}
{\begin{sideways}$|\hat{\mu}_E - \hat{\mu}_A|$\end{sideways}}

& 1 
  & 0.000 & 0.000 & 0.000 & 0.004 & 0.158 & 0.005 & 0.011\\
& 2 
  & 0.000 & 0.001 & 0.005 & 0.136 & 0.209 & 0.110 & 0.167\\
& 3 
  & 0.001 & 0.004 & 0.006 & 0.173 & 0.079 & 0.126 & 0.085\\
& 4 
  & 0.003 & 0.004 & 0.016 & 0.116 & 0.121 & 0.158 & 0.026\\
& 5 
  & 0.002 & 0.016 & 0.094 & 0.096 & 0.004 & 0.039 & 0.030\\
& 6 
  & 0.001 & 0.041 & 0.095 & 0.000 & 0.176 & 0.045 & 0.038\\
& 7 
  & 0.001 & 0.039 & 0.057 & 0.046 & 0.017 & 0.000 & 0.000\\
& 8 
    & 0.006 & 0.075 & 0.002 & 0.156 & 0.016 & 0.000 & 0.000\\
  
  \bottomrule
\end{tabular}\vspace{0.2cm}
\caption{The absolute differences between the trained and expert mean feature expectations. Each column 
 refers to one of the first 7 sensors 
 and each row refers to one of the first 8 distance bins.}\label{tableResults}
\end{table}
\end{enumerate}

\section{Conclusion}
In this paper we proposed using Deep Q-Networks as the refinement step in Inverse Reinforcement Learning approaches. This enabled us to extract the rewards in scenarios with large state spaces such as driving, given expert demonstrations. The aim of this work was to extend the general approach to IRL. Exploring more advanced methods like Maximum Entropy IRL and the support for nonlinear reward functions is currently under investigation.




\small

\begin{thebibliography}{10}
	\providecommand{\url}[1]{#1}
	\csname url@rmstyle\endcsname
	\providecommand{\newblock}{\relax}
	\providecommand{\bibinfo}[2]{#2}
	\providecommand\BIBentrySTDinterwordspacing{\spaceskip=0pt\relax}
	\providecommand\BIBentryALTinterwordstretchfactor{4}
	\providecommand\BIBentryALTinterwordspacing{\spaceskip=\fontdimen2\font plus
		\BIBentryALTinterwordstretchfactor\fontdimen3\font minus
		\fontdimen4\font\relax}
	\providecommand\BIBforeignlanguage[2]{{%
			\expandafter\ifx\csname l@#1\endcsname\relax
			\typeout{** WARNING: IEEEtran.bst: No hyphenation pattern has been}%
			\typeout{** loaded for the language `#1'. Using the pattern for}%
			\typeout{** the default language instead.}%
			\else
			\language=\csname l@#1\endcsname
			\fi
			#2}}
	
	\bibitem{kruse2013human}
	T.~Kruse, A.~K. Pandey, R.~Alami, and A.~Kirsch, ``Human-aware robot
	navigation: A survey,'' \emph{Robotics and Autonomous Systems}, vol.~61,
	no.~12, pp. 1726--1743, 2013.
	
	\bibitem{triebel15spencer}
	R.~Triebel, K.~Arras, R.~Alami, L.~Beyer, S.~Breuers, R.~Chatila, M.~Chetouani,
	D.~Cremers, V.~Evers, M.~Fiore, H.~Hung, O.~A.~I. Ram\'{i}rez, M.~Joosse,
	H.~Khambhaita, T.~Kucner, B.~Leibe, A.~J. Lilienthal, T.~Linder, M.~Lohse,
	M.~Magnusson, B.~Okal, L.~Palmieri, U.~Rafi, M.~van Rooij, and L.~Zhang,
	``Spencer: A socially aware service robot for passenger guidance and help in
	busy airports,'' in \emph{Proc. Field and Service Robotics (FSR)}, 2015.
	
	\bibitem{okallearning}
	B.~Okal and K.~O. Arras, ``Learning socially normative robot navigation
	behaviors with bayesian inverse reinforcement learning,'' in \emph{Robotics
		and Automation (ICRA), 2016 IEEE International Conference on}.\hskip 1em plus
	0.5em minus 0.4em\relax IEEE, 2016, pp. 2889--2895.
	
	\bibitem{werling2010optimal}
	M.~Werling, J.~Ziegler, S.~Kammel, and S.~Thrun, ``Optimal trajectory
	generation for dynamic street scenarios in a frenet frame,'' in
	\emph{Robotics and Automation (ICRA), 2010 IEEE International Conference
		on}.\hskip 1em plus 0.5em minus 0.4em\relax IEEE, 2010, pp. 987--993.
	
	\bibitem{tehrani2014evaluating}
	H.~Tehrani, K.~Muto, K.~Yoneda, and S.~Mita, ``Evaluating human \& computer for
	expressway lane changing,'' in \emph{2014 IEEE Intelligent Vehicles Symposium
		Proceedings}.\hskip 1em plus 0.5em minus 0.4em\relax IEEE, 2014, pp.
	382--387.
	
	\bibitem{tehrani2015general}
	H.~Tehrani, Q.~H. Do, M.~Egawa, K.~Muto, K.~Yoneda, and S.~Mita, ``General
	behavior and motion model for automated lane change,'' in \emph{2015 IEEE
		Intelligent Vehicles Symposium (IV)}.\hskip 1em plus 0.5em minus 0.4em\relax
	IEEE, 2015, pp. 1154--1159.
	
	\bibitem{bojarski2016end}
	M.~Bojarski, D.~Del~Testa, D.~Dworakowski, B.~Firner, B.~Flepp, P.~Goyal, L.~D.
	Jackel, M.~Monfort, U.~Muller, J.~Zhang, \emph{et~al.}, ``End to end learning
	for self-driving cars,'' \emph{arXiv preprint arXiv:1604.07316}, 2016.
	
	\bibitem{ng2000algorithms}
	A.~Y. Ng, S.~J. Russell, \emph{et~al.}, ``Algorithms for inverse reinforcement
	learning.'' in \emph{Icml}, 2000, pp. 663--670.
	
	\bibitem{abbeel2004apprenticeship}
	P.~Abbeel and A.~Y. Ng, ``Apprenticeship learning via inverse reinforcement
	learning,'' in \emph{Proceedings of the twenty-first international conference
		on Machine learning}.\hskip 1em plus 0.5em minus 0.4em\relax ACM, 2004, p.~1.
	
	\bibitem{ramachandran2007bayesian}
	D.~Ramachandran and E.~Amir, ``Bayesian inverse reinforcement learning,''
	\emph{Urbana}, vol.~51, no. 61801, pp. 1--4, 2007.
	
	\bibitem{ziebart2008maximum}
	B.~D. Ziebart, A.~L. Maas, J.~A. Bagnell, and A.~K. Dey, ``Maximum entropy
	inverse reinforcement learning.'' in \emph{AAAI}, 2008, pp. 1433--1438.
	
	\bibitem{ratliff2006maximum}
	N.~D. Ratliff, J.~A. Bagnell, and M.~A. Zinkevich, ``Maximum margin planning,''
	in \emph{Proceedings of the 23rd international conference on Machine
		learning}.\hskip 1em plus 0.5em minus 0.4em\relax ACM, 2006, pp. 729--736.
	
	\bibitem{wulfmeier2015deep}
	M.~Wulfmeier, P.~Ondruska, and I.~Posner, ``Deep inverse reinforcement
	learning,'' \emph{arXiv preprint arXiv:1507.04888}, 2015.
	
	\bibitem{mnih2015human}
	V.~Mnih, K.~Kavukcuoglu, D.~Silver, A.~A. Rusu, J.~Veness, M.~G. Bellemare,
	A.~Graves, M.~Riedmiller, A.~K. Fidjeland, G.~Ostrovski, \emph{et~al.},
	``Human-level control through deep reinforcement learning,'' \emph{Nature},
	vol. 518, no. 7540, pp. 529--533, 2015.
	
	\bibitem{watkins1989learning}
	C.~J. C.~H. Watkins, ``Learning from delayed rewards,'' Ph.D. dissertation,
	King's College, Cambridge, 1989.
	
	\bibitem{watkins1992q}
	C.~J. Watkins and P.~Dayan, ``Q-learning,'' \emph{Machine learning}, vol.~8,
	no. 3-4, pp. 279--292, 1992.
	
	\bibitem{tsitsiklis1997analysis}
	J.~N. Tsitsiklis and B.~Van~Roy, ``An analysis of temporal-difference learning
	with function approximation,'' \emph{IEEE transactions on automatic control},
	vol.~42, no.~5, pp. 674--690, 1997.
	
	\bibitem{lavalle2006planning}
	S.~M. LaValle, \emph{Planning algorithms}.\hskip 1em plus 0.5em minus
	0.4em\relax Cambridge university press, 2006.
	
\end{thebibliography}

\end{document}